\documentclass{article}

\usepackage[preprint]{nips_2018}
\usepackage[utf8]{inputenc}
\usepackage{hyperref}       % hyperlinks
\usepackage{url}            % simple URL typesetting
\usepackage{microtype}
\usepackage{graphicx}
\usepackage{subfigure}
\usepackage{booktabs} % for professional tables
\usepackage{amsmath}
\usepackage{caption}
\usepackage{array}
\usepackage{listings}
\usepackage[title]{appendix}
\DeclareMathOperator{\sign}{sign}

\title{DARCCC: Detecting Adversaries by Reconstruction from Class Conditional Capsules}
\author{Nicholas Frosst, Sara Sabour, Geoffrey Hinton\\
\texttt{\!\{frosst,\,sasabour,\,geoffhinton\}@google.com} \\
Google Brain\\% Toronto, ON, Canada\\
}
\date{August 2018}

\begin{document}

\maketitle

\begin{abstract}
We present a simple technique that allows capsule models to detect adversarial images. In addition to being trained to classify images, the capsule model is trained to reconstruct the images from the pose parameters and identity of the correct top-level capsule. Adversarial images do not look like a typical member of the predicted class and they have much larger reconstruction errors when the reconstruction is produced from the top-level capsule for that class.  We show that setting a threshold on the $l2$ distance between the input image and its reconstruction from the winning capsule is very effective at detecting adversarial images for three different datasets. The same technique works quite well for CNNs that have been trained to reconstruct the image from all or part of the last hidden layer before the softmax.   We then explore a stronger, white-box attack that takes the reconstruction error into account. This attack is able to fool our detection technique but in order to make the model change its prediction to another class, the attack must typically make the "adversarial" image resemble images of the other class. 
\end{abstract}

\section{Introduction}
%Adversarial examples represent a failure mode of neural networks that surprises and intrigues the machine learning community. One reason for this interest is the relative imperceptibility of attacks, i.e. it is possible to make unnoticeable changes to a correctly classified input image so as to fool the classifier into misclassifying the altered image. In a typical classification neural network, understanding what the network `sees' in the input and how it perceives the image besides the output classification is an intractable task. 

\cite{sabour2017} show that the discriminative performance of a capsule network can be improved by adding another network that reconstructs the input image from the pose parameters and the identity of the correct top-level capsule. Derivatives back-propagated through the reconstruction network force the pose parameters of the top-level capsule to capture a lot of information about the image.  A capsule network trained with such a regularizer can output not only a classification, but also a class conditional reconstruction of the input.
%which shows conditions on the understanding of the CapsNet from the image encoded by pose parameters. 
We show that the reconstruction sub-network can be used as a very effective way to detect adversarial attacks: we reconstruct the input from the identity and pose parameters of the winning top-level capsule to verify that the network is perceiving what we expect it to perceive in a typical example of that class. We propose DARCCC which is an attack independent detection technique relying on the difference between the distribution of class reconstruction distances for genuine images vs adversarial images. We extend DARCCC to more standard image classification networks (convolution neural networks) and we show the effectiveness of our detection method against black box attacks and typical white box attacks on three image data-sets; MNIST, Fashion-MNIST and SVHN.

Our detection method can be defeated by a stronger white-box attack that uses a method (R-BIM) that takes the reconstruction error into account and iteratively perturbs the image so as to allow good reconstruction. However, this stronger attack does not produce typical adversarial images that look like the original image but with a small amount of added noise. Instead, in order to make the model classify the image incorrectly, the perturbation to the original image must be substantial and typically leads to an "adversarial" image that actually resembles other images of the target class.  Moreover, for a capsule network, if enough weight is put on the reconstruction error to avoid detection, it is often impossible to change the image in a way that causes the desired misclassification.  

\begin{figure}[t]
  \centering
  \includegraphics[width=0.5\linewidth]{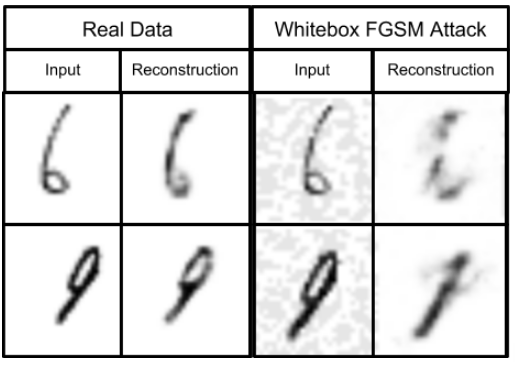}
  \caption{The reconstructions from the predicted class pose parameters of a trained capsule network for real data and successfully adversary of same data with a target class of '1'. The reconstructions from adversarial data resemble a '1' more than the input image.}
  \label{fig:adversarial_vs_real_reconstructions}
\end{figure}

\section{Background}
\cite{biggio2013evasion} introduced the adversaries for Machine Learning systems. Imperceptible adversarial images for Deep Neural Networks were introduced by \cite{szegedy2013} where they used a second order optimizer. Fast Gradient Sign method (\cite{goodfellow2014}) showed that by taking an $\epsilon$ step in the direction of the gradient, $-\epsilon \sign \bigl( \nabla_X J(X, \text{label}) \bigr) $, one can change the label of the input image $X$. Such adversarial attacks which have access to the attacked model are called ``white box'' attacks. \cite{goodfellow2014} also showed the effectiveness of ``black box'' attacks where the adversarial images for a model is used to attack another model. 
Basic Iterative Method (\cite{kurakin2016}) takes multiple $\alpha$-wide FGSM steps in the $\epsilon$ ball of the original image.

\cite{kurakin2018} provides an overview of the oscillating surge of attacks and defenses. Recently several generative approaches are proposed (\cite{samangouei2018,ilyas2017,meng2017magnet}) which assume adversarial images does not exist on the input image manifold. \cite{carlini2017adversarial} depicts failure of such adversarial detection techniques. \cite{jetley2018} and \cite{gilmer2018} investigate relation of adversarial images to the accuracy of the model and to the input manifold. Since our method conditions on the prediction of the model for generating an image it does not depend on this assumptions. Most recently, \cite{schott2018} investigated effectiveness of a class conditional generative model as a defense mechanism for MNIST digits. Our method in comparison, does not increase the computational overhead of the classification.

\cite{sabour2015} shows that adversaries exist for a network with random weights. Therefore, susceptibility to adversarial attacks is not caused by learning and the convolution neural network architectures are inherently fragile. Capsule networks (\cite{sabour2017,hinton2018}) are a new neural network architecture where neurons activate based on agreement of incoming vectors and defer architecturally from Convolutional neural networks. This new architecture has been proven to be more robust to white box attacks while being as weak as CNNs in defending  black box attacks. In this work we address this shortcoming by introducing an adversarial detection mechanism based on reconstruction sub-network of CapsNets. Furthermore, we extend this technique to typical CNNs.
\section{Method}
\begin{figure}[t]
  \centering
  \includegraphics[width=\linewidth]{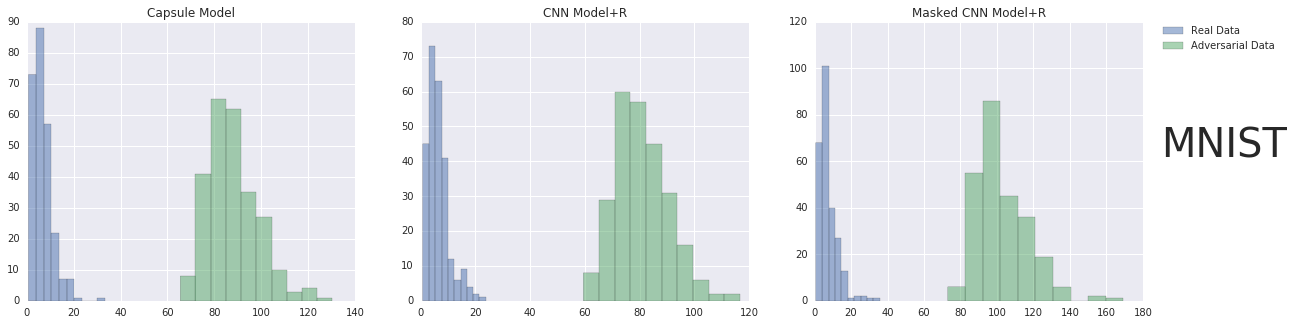}
  \caption{The histogram of $l2$ distances between the reconstruction and the input for each model, for real and adversarial data for MNIST. We used FGSM with $\epsilon=0.3$ to create the attacks.}
  \label{fig:hist}
\end{figure}

The reconstruction network of the CapsNet proposed in \cite{sabour2017} takes in the pose parameters of all the class capsules and mask all values to 0 except for the pose parameters of the predicted class. During training they optimize the $l2$ distance of input image and reconstruction along side the classification loss. We use the same reconstruction network for detecting adversarial attacks by measuring the euclidean distance between the input and a prediction reconstruction. Fig.~\ref{fig:hist} shows a sample histogram of distances for natural images vs adversarial images. We leverage the difference between the two mentioned distributions and propose DARCCC for detecting adversaries based on the reconstruction from classification. DARCCC distinguishes adversaries by thresholding images based on their reconstruction distance. Fig.~\ref{fig:adversarial_vs_real_reconstructions} shows the reconstructions from real and adversarial data; the deviation of the adversarial reconstructions from input image motivates this approach.

Although the system above is designed for informative pose parameters of the CapsNet, the strategy can be extended beyond CapsNets. We create a similar architecture,  ``Masked CNN+R'', by using a standard CNN and dividing the penultimate hidden layer into groups corresponding each class. The sum of each neuron group serves as the logit for that particular class and the group itself is passed to the reconstruction sub-network via the same masking operation used by \cite{sabour2017}. We also study the role of class conditional reconstruction by omitting the masking and experimenting with a typical ``CNN+R'' model whose entire penultimate layer is used for reconstruction.

\subsection{Detection threshold}
We find the threshold for DARCCC based on the expected distance between a validation input image and its reconstruction. If the distance between the input and the reconstruction is above the chosen threshold DARCCC classifies the data as adversarial. Choosing the threshold poses a trade off between false positive and false negative detection rates. Therefore, it should be chosen based on the assumed likelihood of the system being attacked. Such a trade off is discussed by \cite{gilmer2018}. In our experiments we don't tune this parameter to attacks and set it as the 95th percentile of validation distances. This means our false positive rate on real validation data is 5\%.
\section{Experiments}
The three models, Capsule, CNN+R, and Masked CNN+R, are designed to have the same number of parameters. Fig.~\ref{fig:arch} shows the architecture we use for each one. For our experiments, all were trained with the same Adam optimizer and for the same number of epochs. We did not do an exhaustive parameters search on these models, instead we chose hyper-parameters that allowed each model to perform roughly equivalently on the test sets. Tab.~\ref{tab:acc} shows the test accuracy of these trained models on the three datasets in our experiments, MNIST \citep{lecun1998}, FashionMNIST \citep{xiao2017}, and SVHN \citep{netzer2011}.  
\begin{table}[b]
\caption{The test accuracy of each model for each dataset in our experiments.}
\label{tab:acc}
\centering
\begin{tabular}{|l|l|l|l|l|}
\hline
Dataset      & Capsule Model & CNN+R Model & Masked CNN+R Model   \\ \hline
MNIST        & 0.994         & 0.993     & 0.994              \\ \hline
FashionMNIST & 0.904         & 0.905     & 0.907              \\ \hline
SVHN         & 0.890         & 0.907     & 0.905              \\ \hline
\end{tabular}
\end{table}
\subsection{Black box adversarial attack detection}
\begin{figure}
  \centering
  \includegraphics[width=\linewidth]{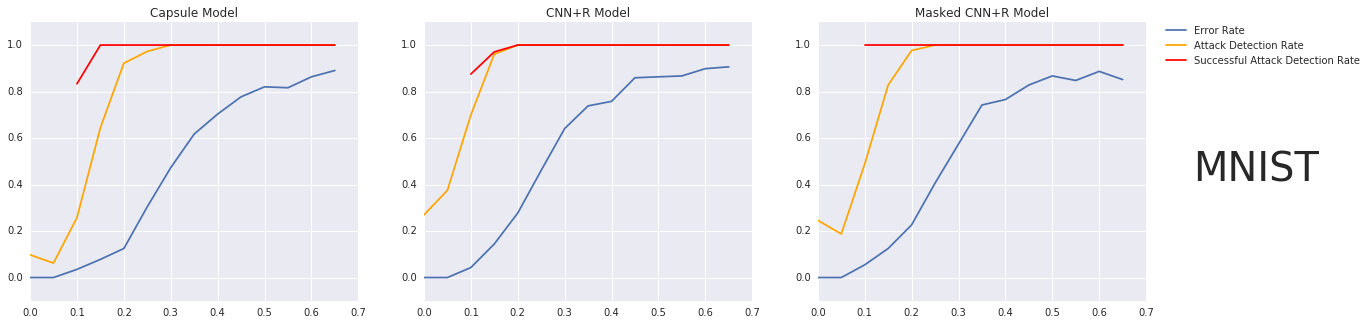}
  \caption{DARCCC detection rates and black box FGSM attack error rates for different $\epsilon$.}
  \label{fig:MNIST_blackbox}
\end{figure}
To test DARCCC on a black box attack, we trained a standard CNN with two layers of convolutions and 2 hidden layers without the aforementioned reconstruction network, and used it to create adversarial attacks using the Fast Gradient Sign Method. Fig.~\ref{fig:MNIST_blackbox} plots the error rate, the attack detection rate, and the successful attack detection rate for each of the 3 models over varying $\epsilon$. For all 3 models, DARCCC not only accurately detects the successful attacks (Successful Attack Detection Rate, attacks which changed the networks classification), it detects perturbations regardless of if they changed the networks classification as well (Attack Detection Rate).

\subsection{White box adversarial attack detection}
\begin{figure}
  \centering
  \includegraphics[width=\linewidth]{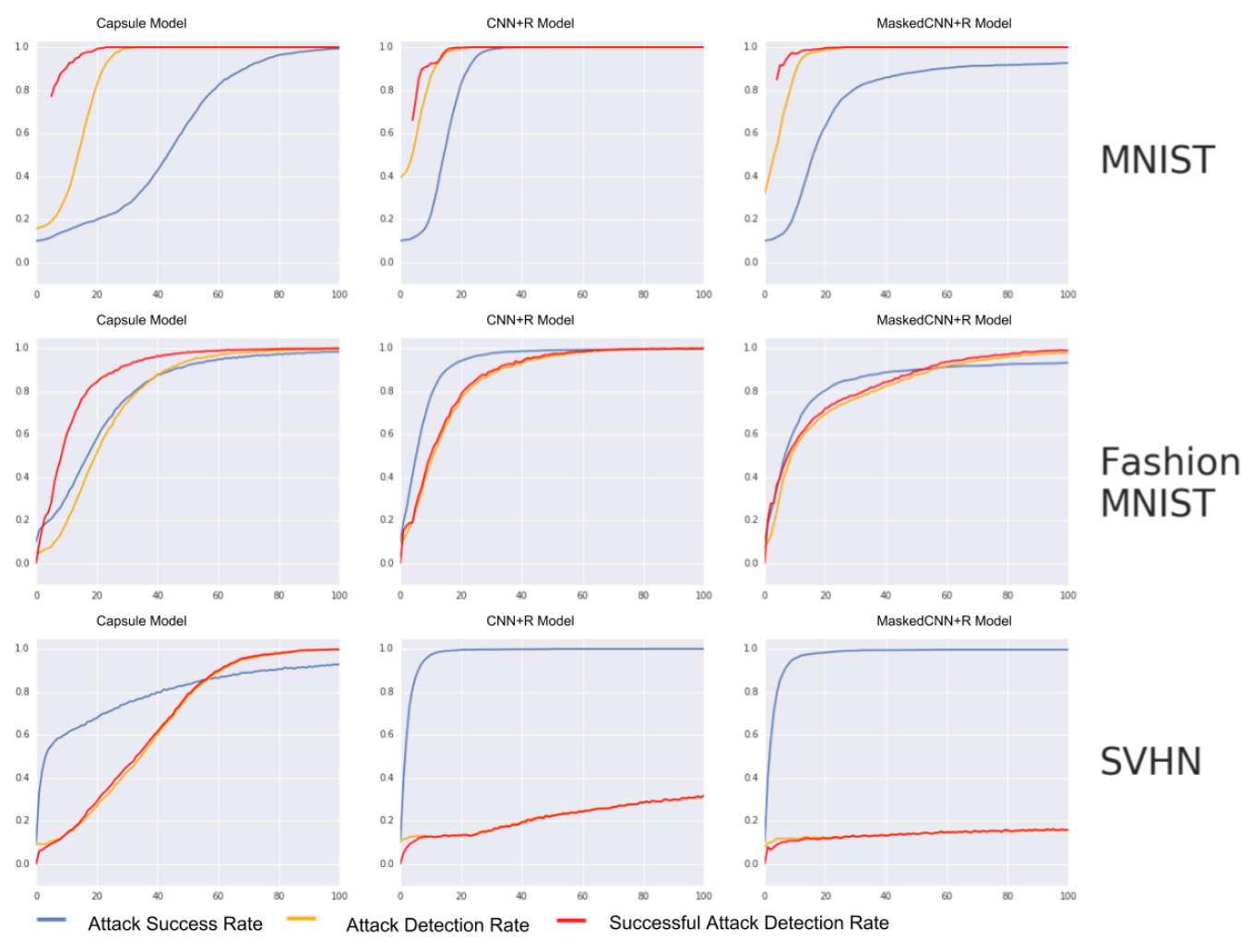}
  \caption{DARCCC detection rate and white box BIM attack success rate at different steps.}
  \label{fig:standard_attack}  
\end{figure}

We tested DARCCC against white box Basic Iterative Method adversarial attacks targeting each class. We use $\alpha=0.01$ and $\epsilon=\alpha \times N_{\text steps}$. We also clipped the result to be between 0 and 1. The success rate of the attack (flipping the classification to the target class), the attack detection rate (whether the image is tampered with), and the successful attack detection rates (detecting images whose prediction has flipped) are plotted in Fig.~\ref{fig:standard_attack} for all three models and for the 3 data sets as a function of the number of steps. For all models, DARCCC is able to detect attacks to some degree for Fashion MNIST and MNIST, but on the capsule model it is able to detect adversaries on SVHN as well.

\subsubsection{Reconstructive BIM attack}
\begin{figure}
  \centering
  \includegraphics[width=\linewidth]{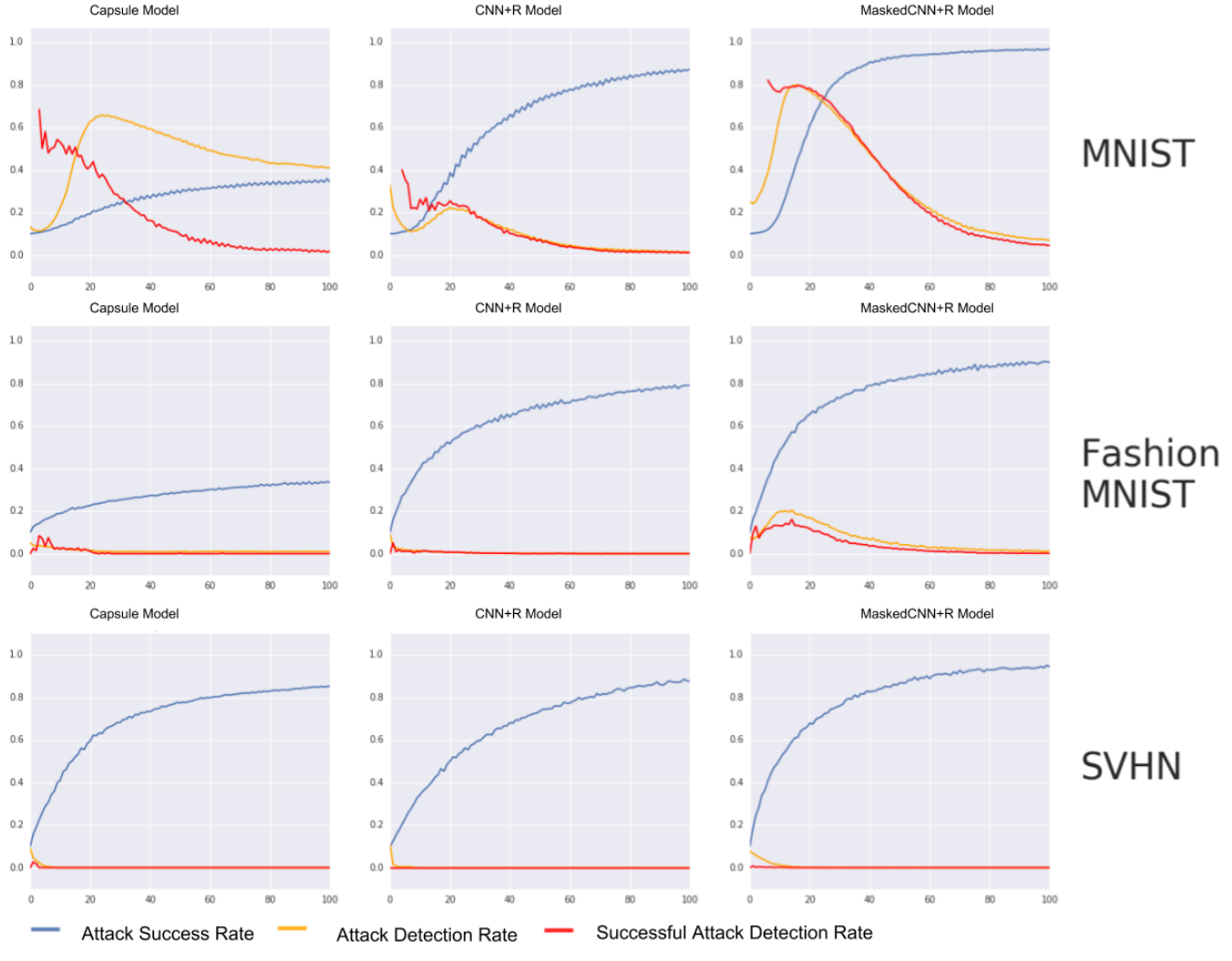}
  \caption{R-BIM attack success rate and DARCCC detection rate for different steps. }
  \label{fig:Recons_attack}
\end{figure}

Targeted BIM takes gradient steps to maximize the classification probability of the target class. Since the reconstruction distance is also differentiable we modify BIM into R-BIM which additionally minimizes the reconstruction distance. R-BIM is designed specifically to break DARCCC.  
\begin{table}
  \centering
\begin{tabular}{m{4.5cm}m{8cm}}
  Seed Image & \includegraphics[width=\linewidth]{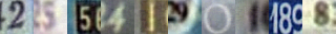}\\ \hline
  R-BIM to 0 for Capsule & \includegraphics[width=\linewidth]{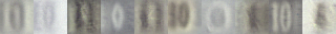}\\ \hline
  R-BIM to 0 for Masked-CNN+R & \includegraphics[width=\linewidth]{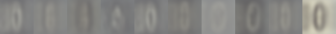}\\ \hline
  R-BIM to 0 for CNN+R & \includegraphics[width=\linewidth]{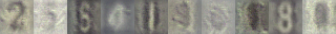}\\
  \end{tabular}
  \captionof{figure}{Top row are the starting SVHN images which are R-BIM attacked to `0'. The next rows show the successful generated adversary for Capsule, Masked-CNN+R and CNN+R model.}
  \label{fig:Recons_attack_examples}
\end{table}
Fig.~\ref{fig:Recons_attack_examples} visualizes the initial input and the result of 100 steps of R-BIM with a target class of `0' for 10 random SVHN images. We see that indeed several of the crafted examples look like `0's. Effectively they are not adversarial images at all since they resemble their predicted class to the human eye. This implies that the gradient is aligned with the true data manifold. Similar visualizations for MNIST and fashion-MNIST can be found in the appendix. For Fashion-MNIST only the capsule model attacks resemble true images from the target class.
We still report the same detection rate plots as above in Fig.~\ref{fig:Recons_attack} for R-BIM. Notably R-BIM is significantly less successful than a standard BIM attack in changing the classification. The capsule model in particular exhibits significant resilience to this attack.

\section{Discussion}
We have presented DARCCC, a simple architectural extension that enables adversarial attack detection. DARCCC  notably relies on a similarity metric between the reconstruction and the input. This metric is required both during training in order to train the reconstruction network and during test time in order to flag adversarial examples. In the 3 data sets we have evaluated, the distance between examples roughly correlates with semantic similarity. This however is not the case for images in more complex data set such as Cifar10 or ImageNet, in which two images may be similar in terms of content or look, but have significant $l2$ distance.  This issue will need to be resolved for this method to scale up to more complex problems, and offers a promising avenue for future research. 

Notably DARCCC does not rely on a specific predefined adversarial attack. We have shown that by reconstructing the input from the internal class-conditional representation, our system is able to accurately detect black box and white box FGSM and BIM attacks. Of the three models we explored, we showed that the capsule model was the best fitted for this task, and was able to detect adversarial examples with greater accuracy on all the data-sets we explored. We then proposed a new, stronger attack to beat our defense - the Reconstructive BIM attack - in which the adversary optimizes not only the classification loss but also the reconstruction loss. We showed that this attack was less successful than a standard attack, and in particular the capsule model showed great resilience. For more complicated data-sets such as SVHN we showed that the detection method was not able to detect the strong adversarial attacks, but when we visualized the perturbed images they typically appeared to be on the true data manifold and from the target class, so they lacked the paradoxical property of typical adversarial attacks.

{\Large{A}}{{CKNOWLEDGEMENT}}.
We thank Mohammad Norouzi and Nicolas Papernot for their feedback, insight and support.

\clearpage
\bibliography{bib}
\bibliographystyle{plainnat}

\begin{appendices}
\renewcommand\thefigure{\thesection.\arabic{figure}}
\renewcommand\thetable{\thesection.\arabic{table}}

\setcounter{figure}{0} 
\setcounter{table}{0} 
\clearpage
\section{Architecture}
\begin{figure}[h]
  \centering
  \includegraphics[width=\linewidth]{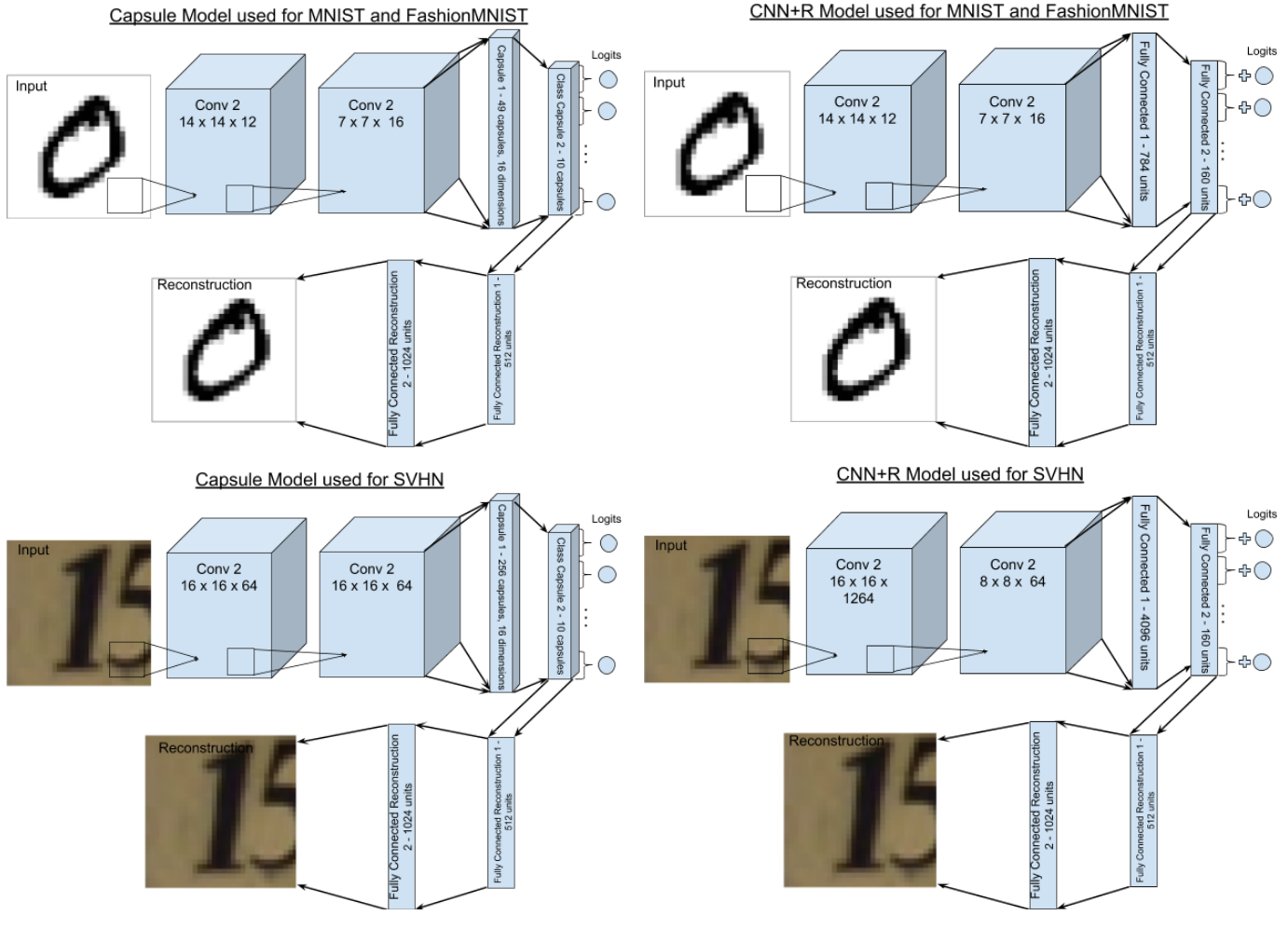}
  \caption{The architecture for the CNN+R and Capsule model used for our experiments on MNIST, Fashion-MNIST, and SVHN.}
   \label{fig:arch}
\end{figure}

Fig.~\ref{fig:arch} shows the architecture of the Capsule network and the CNN+R model used for experiments on MNIST, Fashion-MNIST and SVHN. MNSIT and Fashion-MNIST have exactly same architectures while for SVHN experiments we use larger models. Note that the only difference between the CNN+R and the Masked CNN+R is the masking procedure on the input to the reconstruction network based on the predicted class. All three models have the same number of parameters for each dataset.

\clearpage
\section{Histogram of distances}
\begin{figure}[h]
  \centering
  \includegraphics[width=\linewidth]{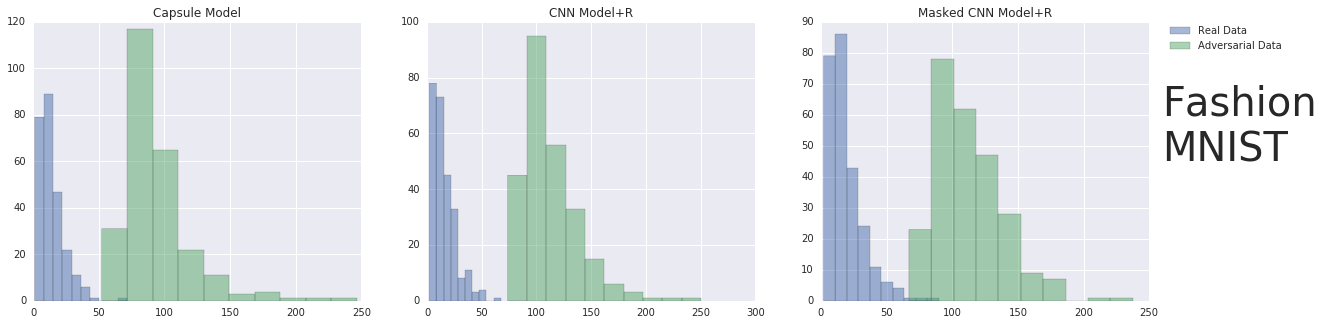}
  \caption{FashionMNIST - The histogram of $l2$ distances between the reconstruction and the input for each model, for real and adversarial data.}
  \label{fig:hist_fashion_mnist}
\end{figure}

\begin{figure}[h]
  \centering
  \includegraphics[width=\linewidth]{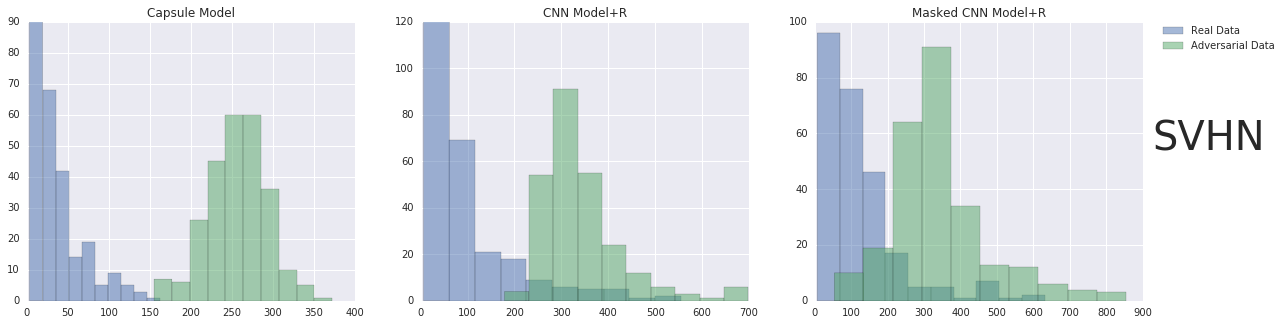}
  \caption{SVHN - The histogram of $l2$ distances between the reconstruction and the input for each model, for real and adversarial data.}
  \label{fig:hist_svhn}
\end{figure}

Fig.~\ref{fig:hist_fashion_mnist} and Fig.~\ref{fig:hist_svhn} visualize the histogram of euclidean distances for real Fashion-MNIST and SVHN validation images (blue) vs the white box FGSM with $\epsilon=0.3$ adversarial images (green) as a proof of concept and motivation. We do not factor the distribution of adversarial distances for picking DARCCC threshold. The threshold is solely based on the validation distances.

\section{Black box attacks}
\begin{figure}[h]
  \centering
  \includegraphics[width=\linewidth]{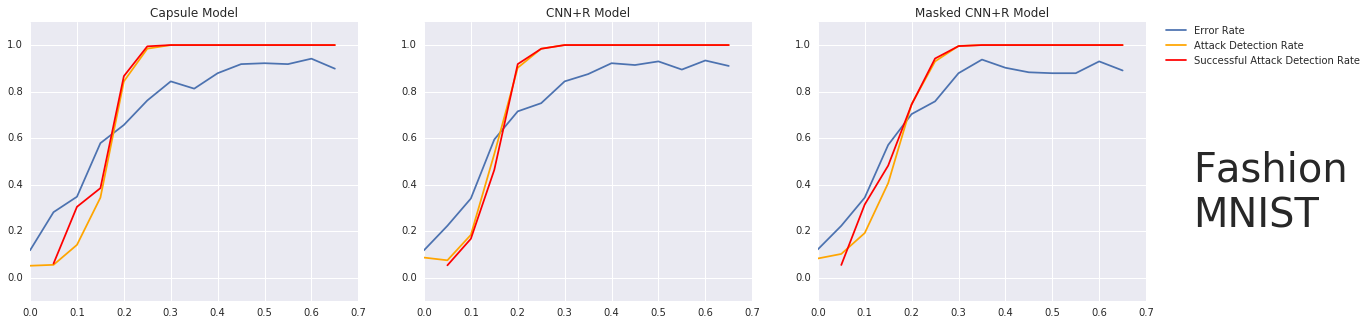}
  \includegraphics[width=\linewidth]{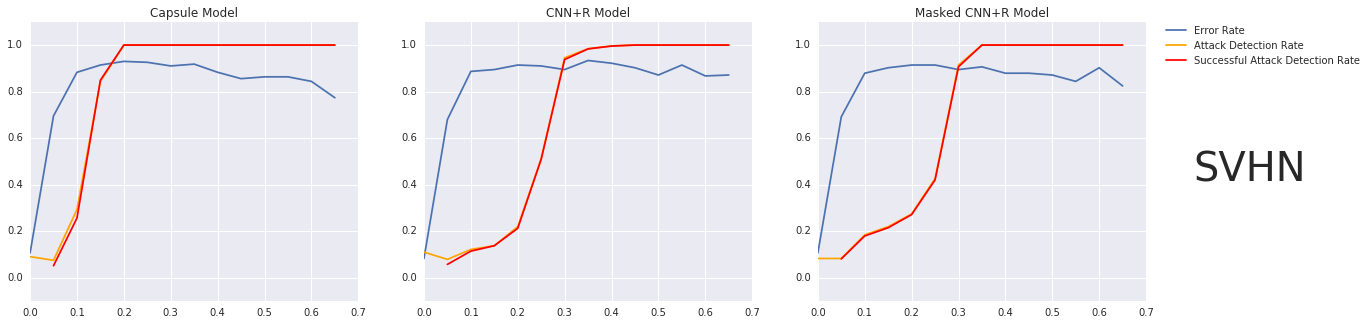}
  \label{fig:blackbox_fgsm_detection_rate}
\end{figure}

\clearpage
\section{R-BIM attack samples}

\begin{figure}[h]
  \centering
  \textbf{Capsule}
  \includegraphics[width=\linewidth]{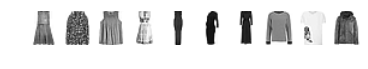}
  \includegraphics[width=\linewidth]{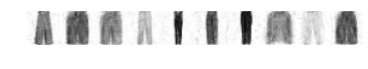}
  \textbf{MaskedCNN+R }
  \includegraphics[width=\linewidth]{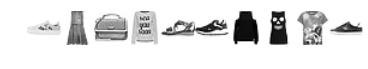}
  \includegraphics[width=\linewidth]{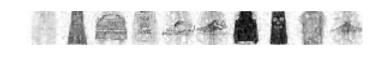}
  \textbf{CNN+R}
  \includegraphics[width=\linewidth]{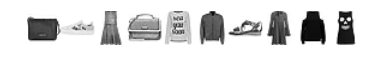}
  \includegraphics[width=\linewidth]{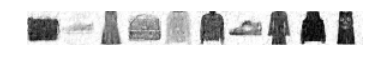}
  \label{fig:fashionmnist_recons_attack}
  \caption{Initial image (top) and Reconstruction BIM (100 steps of 0.01) perturbed image (bottom) for each model.  These images where successful attacks targetting the pants class. They were chosen at random from successful attacks. Only the Capsule Model Attacks resemble images from the target class (pants). In case of CapsNet there were no shoes successfully flipped into pants. }
\end{figure}

\begin{figure}[h]
  \centering
  \textbf{Capsule}
  \includegraphics[width=\linewidth]{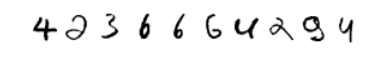}
  \includegraphics[width=\linewidth]{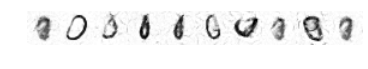}
  \textbf{MaskedCNN+R }
  \includegraphics[width=\linewidth]{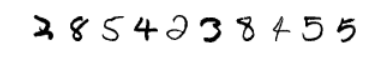}
  \includegraphics[width=\linewidth]{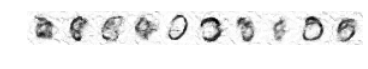}
  \textbf{CNN+R}
  \includegraphics[width=\linewidth]{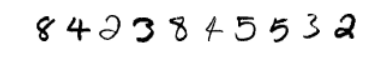}
  \includegraphics[width=\linewidth]{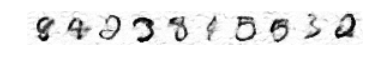}
  \label{fig:fashionmnist_recons_attack}
  \caption{Initial image (top) and Reconstruction BIM (100 steps of 0.01) perturbed image (bottom) for each model.  These images are successful attacks targetting class `0'. They were chosen at random from successful attacks. }
\end{figure}

\clearpage
\section{All Class Capsule Reconstructions}

\begin{figure}[h]
  \centering
  \includegraphics[width=\linewidth]{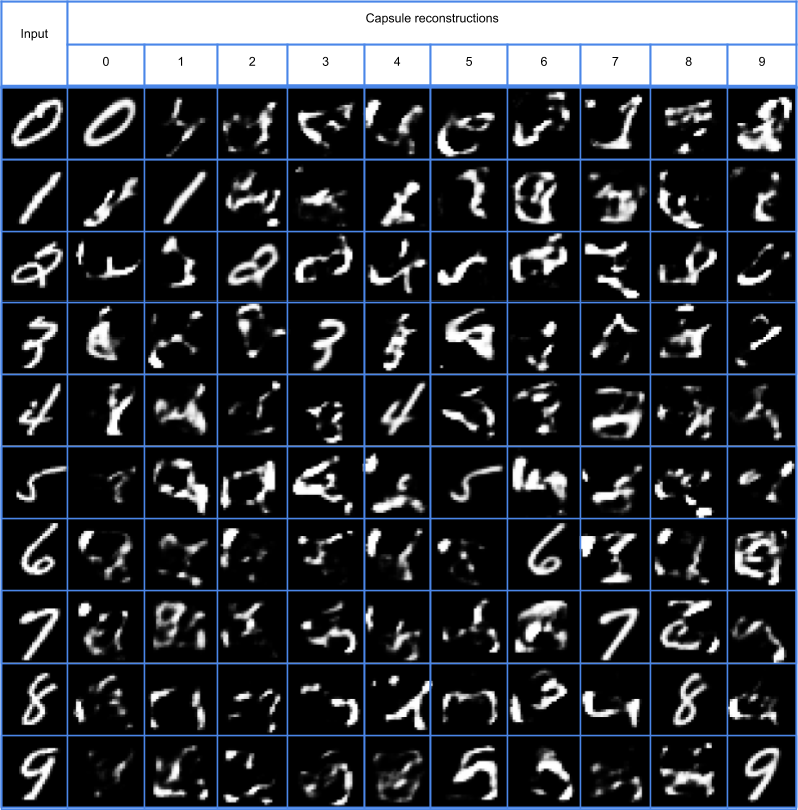}
  \caption{Input and Reconstructions from each class capsule - This graphic explains the motivation for using the distance between the input and reconstruction as an adversarial defense; for the most part only the reconstructions from the correct capsule are sensible.}
\end{figure}
\begin{figure}[h]
  \centering
  \includegraphics[width=\linewidth]{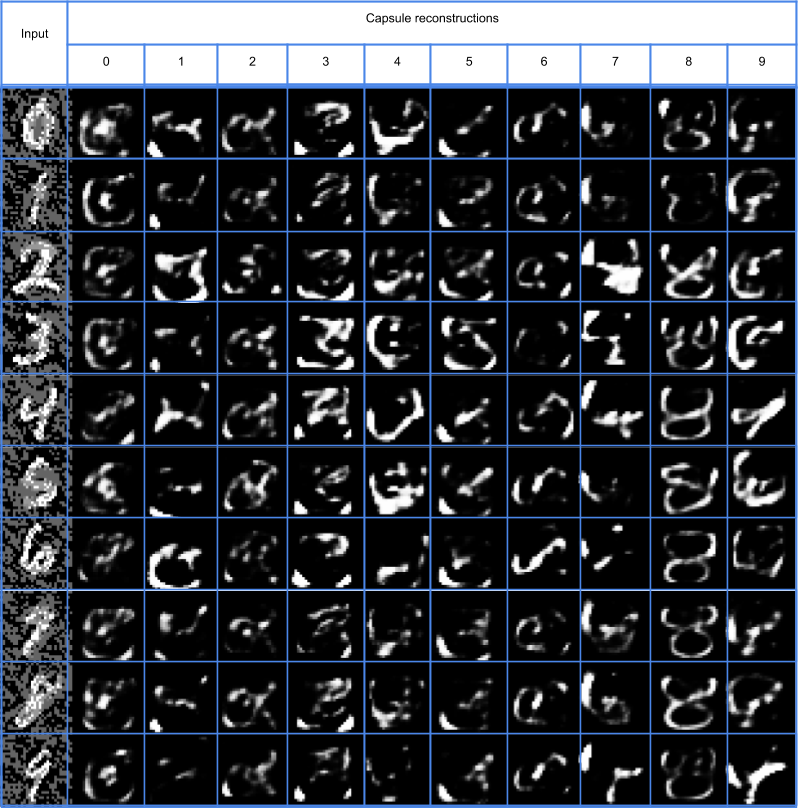}
  \caption{Input and Reconstructions from each class capsule for FGSM attacks - we can see that the reconstructions do not look sensible for any class, neither the predicted class nor the labeled class capsule reconstruction resemble to the input.}
\end{figure}

\end{appendices}

\end{document}